\documentclass[conference]{IEEEtran}
\IEEEoverridecommandlockouts
\usepackage{cite}
\usepackage{amsmath,amssymb,amsfonts}
\usepackage{algorithmic}
\usepackage{graphicx}
\usepackage{textcomp}
\usepackage{xcolor}
\usepackage{pgf}
\usepackage{subcaption}
\def\BibTeX{{\rm B\kern-.05em{\sc i\kern-.025em b}\kern-.08em
    T\kern-.1667em\lower.7ex\hbox{E}\kern-.125emX}}
\begin{document}

\title{On the Fairness of Generative Adversarial Networks (GANs)\\

}

\author{\IEEEauthorblockN{Patrik Joslin Kenfack\IEEEauthorrefmark{1},
Daniil Dmitrievich Arapov\IEEEauthorrefmark{2},
Rasheed Hussain\IEEEauthorrefmark{3},
S.M. Ahsan Kazmi\IEEEauthorrefmark{4}, and
Adil Khan\IEEEauthorrefmark{5}, 
%Author 6\IEEEauthorrefmark{6}
\\ 
%\IEEEauthorblockN{Innopolis University,  Innopolis, Russia. }
\IEEEauthorblockN{ \{\IEEEauthorrefmark{1}p.kenfack, \IEEEauthorrefmark{2}d.arapov\}@innopolis.university} 
\IEEEauthorblockN{\{\IEEEauthorrefmark{3}r.hussain, \IEEEauthorrefmark{4}a.kazmi,\IEEEauthorrefmark{5}a.khan\}@innopolis.ru}  
\IEEEauthorblockN{Innopolis University,  Innopolis, Russia. }
 }}

\maketitle

\begin{abstract}
Generative adversarial networks (GANs) are one of the greatest advances in AI in recent years. With their ability to directly learn the probability distribution of data, and then sample synthetic realistic data. Many applications have emerged, using GANs to solve classical problems in machine learning, such as data augmentation, class imbalance problems, and fair representation learning. In this paper, we analyze and highlight fairness concerns of GANs. In this regard, we show empirically that GANs models may inherently prefer certain groups during the training process and therefore they're not able to homogeneously generate data from different groups during the testing phase. Furthermore, we propose solutions to solve this issue by conditioning the GAN model towards samples' group or using ensemble method (boosting) to allow the GAN model to leverage distributed structure of data during the training phase and generate groups at equal rate during the testing phase.       
\end{abstract}


\section{Introduction}
Machine learning (ML) aims at reproducing human intelligence, for example by learning patterns in data and trying to generalize them. When data are labelled the training process is supervised, i.e. each training instance has a label and the goal of the machine learning model is to learn patterns from these data and predict labels of new instances. When labels are not provided, the training process is said to be unsupervised and the model will try to discover patterns in data. ML models are widely used nowadays to solve different tasks, such as pattern recognition \cite{liu2017survey}, natural language processing \cite{vaswani2017attention}, anomaly detection \cite{rivera2020anomaly}, action recognition\cite{gavrilin2019across, sozykin2018multi, khan2010accelerometer}, scene classification\cite{protasov2018using}, etc.
However, in general the success of these models depend on features used into the training process \cite{bengio2013representation, khan2020post}. Although these features might be provided via features engineering, which requires a lot of work and in certain domains difficult to extract manually, the need for the ML models to extract meaningful features from the data by themselves has become a hot topic within the Artificial Intelligence (AI) community \cite{bengio2013representation, liu2017survey}. This is where \textit{representation learning} (typically Deep Leaning) come in. Basically, the goal is to identify and extract useful information from data to facilitate or improve the predictions, or even generate new data. Generative model is one of the most promising approach in this direction. The main idea behind generative model, is to capture the underlying distribution of the data in order to be able to sample most realistic examples from that learned distribution. 

Among the most popular generative approaches, we have Variational Auto Ecoders (VAEs) \cite{kingma2013auto, rezende2014stochastic} that consist in two neural networks: an encoder and a decoder trained jointly. The encoder tries to encode the prior distribution of the data into a latent space. The decoder tries to reconstruct original examples using examples sampled from the learned distribution with the goal to minimize the reconstruction loss. Although VAEs are simple to train, they suffer from the assumption of an approximate posterior distribution of the observed data. Thus, this makes VAEs less efficient in generating data that most reflect the real distribution of the observed data (e.g poor images quality). Generative Adversarial Networks (GANs) \cite{goodfellow2014generative} solve the shortcoming of VAEs by trying to identify the real distribution of the observed data  using an approach inspired by the Game Theory, the goal to find Nash equilibrium between the two networks, Generator and Discriminator. The generator generates most realistic examples as possible in order to deceive the discriminator, which the aim is to distinguish between the real and the generated (fake) data. Generator and discriminator are trained in competition until Nash equilibrium is reached, which means that the generator was able to generate data similar to the actual observed data such that the discriminator wasn't able to make the difference between them.
Indeed, GANs and its different variants are considered as the most interesting breakthrough in AI for the last decade, because of its outstanding data generation capacity, and and its ability to generate diverse types of data. i.e it allows to generate most realistic images \cite{karras2017progressive}, natural language generation \cite{guo2018long}, music generation \cite{mogren2016c},  medical data generation \cite{esteban2017real}, fair representation learning \cite{madras2018learning} etc.

More importantly, the wide use of AI in different domain in real life has shown that ML models might exhibit  unintended behaviour such as learning biases in data, even amplified them or generate new biases. i.e a model trained using data containing sensitive attributes such gender with disproportionate representation, will tends to be biased toward one groups \cite{dwork2012fairness}. Thus, fairness has become another important research focus within the ML community. The fairness of important technologies such as GANs must be particularly analysed. In general, dataset may contain different groups of data with different distributions, so it is important to analyse and ensure that rather than prioritizing a single group, the generator of GANs models is capable of generating examples of different groups at an equal rate. Several fairness notions have proposed in the literature. Among them, \textit{statistical parity} and \textit{individual fairness} \cite{dwork2012fairness} are widely used. Individual fairness requires that similar individual must receive the same outcome, while statistical parity requires that classifier predictions outcome must be independent to the protected attributes. Bias in data can exist in many shapes and forms, some of which can lead to unfairness in different downstream learning tasks \cite{mehrabi2019survey}. Representation bias is a common form of bias that leads to unfairness decisions, it happens from the way we define and sample from a population. i.e. if certain groups in the population are not represented or underrepresented.  

In paper this we extensively analyse GANs models with respect to their ability to generate examples from different groups present in dataset at an equal rate. To do that, we first present experiments that show that is in most cases, GANs are unable to generate example from different groups and always converge to a single group distribution. 

The key contributions of this paper are the following:
\begin{itemize}
    \item We highlight and demonstrate the shortcomings of GANs in reproducing data from different groups, which may lead to fairness concerns i.e generate biased synthetic data.
     \item Point out the negative impact of inherent biases existing in GANs generation process, on downstream tasks such as classification.
    \item We propose a GAN boosting strategy to generate data samples from different groups at equal rate, by creating diverse ensemble of generators. 
\end{itemize}

The rest of this paper in organised as follows: Section \ref{sec:overview} introduces different GANs models and the theory behind. Section \ref{sec:results} provides the experiment results, that highlight the inability of GANs to generate examples belonging to different group distributions even thought they are were well balanced in the dataset.  Section \ref{sec:solution} presents possible solutions and research directions to fix this issue and finally the discussions and the conclusion. 

\section{Overview of generative adversarial networks}
\label{sec:overview}
Inspired by the game theory, GANs model consist in two neural networks (generator and discriminator) trained in adversarial manner. The generator's (G) goal is to learn the underlying distribution ($p_{g}$) of the dataset by sampling as most realistic examples as possible such the discriminator (D), which the goal is to distinguish between generated examples and real examples, will not be able to make to difference between examples $G(z)$ sampled from $G$ using the random noise vector $z$ and real examples $x$. Thus, $G$ and $D$ play a minimax game with value function $V(G, D)$:

\begin{equation} 
\begin{split}
\begin{aligned}
    \underset{G}{\mathrm{min}} \: \underset{D}{\mathrm{max} }\: V(D, G) = \; & \mathbb{E}_{x \sim p_{data}(x)}[\mathrm{log} \: D(x)]\\ 
    &+ \mathbb{E}_{z \sim p_{z}}[1 - \mathrm{log}\: D(G(z))]
\end{aligned}
\end{split}
\end{equation}

The parameters in G and D will be updated simultaneously during the training until $D(G(z)) = 0.5$, which means that the global optimal solution in achieved and D cannot identify the different between these two distributions. However, this original GANs model has presented some limitations and several extended variants have been proposed. Among these limitations, the problem of mode collapse is the one that has attracted the most attention. This problem occurs when the model can only fit a few modes of the data distribution, while ignoring the majority of them.   In this regards, Arjovsky et. al \cite{arjovsky2017wasserstein} proposed Wasserstein GAN (WGAN) to improve stability of traditional GANs and resolve problems like mode collapse.

The original GANs used Multi Layers Perceptron to train the generator, since Convolutional Neural Networks (CNN) have shown the most interesting results in extracting feature from images, \cite{radford2015unsupervised} proposed Deep Conditional Generative Adversarial Networks (DCGANs). Although DCGANs is able to learn good representation of images and provides great performances in image generation tasks, it suffers from the fact that the generator can collapse in training mode, and thus produces limited varieties of samples \cite{radford2015unsupervised}. Mirza and Osindero \cite{mirza2014conditional} proposed Conditional Adversarial Networks (CGANs), which extends GANs by adding extra information $c$ in both the generator and the discriminator, such that the generator could generate examples constrained by $c$ e.g. class conditional samples. $c$ can be class label, or any kind of auxiliary information. Therefore, the objective function is defined as for GANs as follow:

\begin{equation} 
\begin{split}
\begin{aligned}
    \underset{G}{\mathrm{min}} \: \underset{D}{\mathrm{max} }\: V(D, G) = \; & \mathbb{E}_{x \sim p_{data}(x)}[\mathrm{log} \: D(x|c)]\\ 
    &+ \mathbb{E}_{z \sim p_{z}}[1 - \mathrm{log}\: D(G(z|c))]
\end{aligned}
\end{split}
\end{equation}

It has been shown adding side information can significantly improve the quality of generated samples.
Several extensions of CGANs have been proposed for diverse purpose, for instance Auxiliary GAN  (ACGAN) \cite{odena2017conditional} in which the extra information $c$ is not added to the discriminator but instead the discrimination perform additional tasks such as predicting the side information $c$ i.e outputs the class label of training data, therefore ACGAN use modified version of the loss function to increase the predictive performance for the class label.  

Stacked GANs (SGAN) \cite{huang2017stacked} uses a particular architecture, in which several encoders (DNN), generators and discriminators are put together.  Encoders are stacked and pre-trained in a bottom-up manner. The input image is passed through the first encoder, which output a high-level representation and pass it to next encoder and so on. The uppermost encoder will output the class label. Provided with pre-trained encoder, generators are stacked and trained independently and jointly \footnote{Samples generated by a preview generator is used as input of the next generator} in top-down manner. Each generator takes as inputs  high level representation (from the encoder) and a noise vector and outputs lower-level representation. At each level, one discriminator distinguishes generated representation from "real" representation. The main advantage of SGAN is that it introduces intermediate supervision using representation discriminator at different representation level.   

%Splitting GAN, .  

\section{The shortcomings of GANs in generating data from different group}
\label{sec:results}
In the section we discuss applicability of GANs model in generating data in a settings where the training  data contains different groups with balance or imbalanced representation. We first describe the experimental setup and then present the results on different dataset. 

\subsection{Experimental setup}
%Here we present the experimental setup. i.e. datset used, resources, GANs models implemented, and  the dataset were splitted
This study focuses on two widely used dataset: The MNIST \cite{lecun1998gradient} dataset, which contains 70,000 labelled examples of 28 x 28 images of hand-written digits, and the SVHN \cite{netzer2011reading} dataset, which contains 73,257 labelled examples (10 classes) of 32 x 32 colored images of house number digits. CelibA is a large-scale and diverse face attribute dataset, which contains 202,599 images of 10,177 identities.
\begin{figure}[!]
    \centering
    \includegraphics[width=0.7\columnwidth]{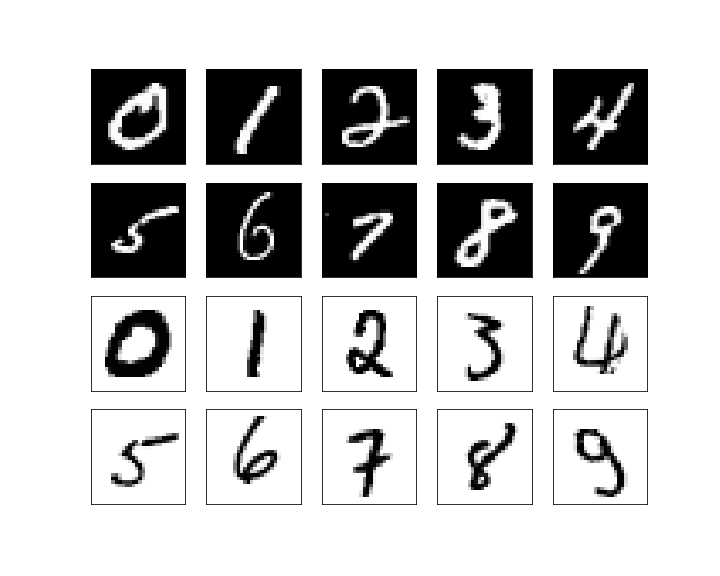}
    \caption{Example of 2 groups created from MNIST dataset. \textit{White digits} having white background and \textit{black digits} with black background}
    \label{fig:mnist_snapshot}
\end{figure}

\begin{figure}[!]
    \centering
    \includegraphics[width=0.7\columnwidth]{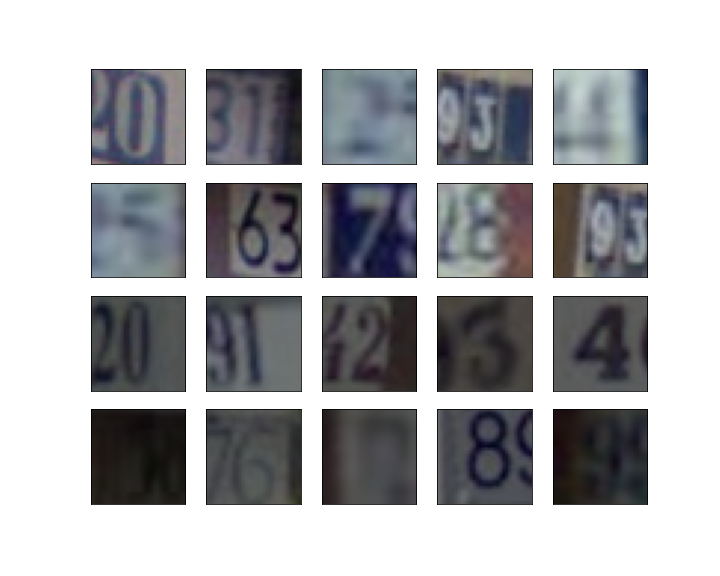}
    \caption{Example of groups created from SVHN dataset. First two rows show samples from the first group (normal samples) and last two rows the second groups (darkened)}
    \label{fig:snvh_snapshot}
\end{figure}

To create different groups, in MNIST dataset, we artificially inverted colors of certain images from black to white (Fig. \ref{fig:mnist_snapshot}) with different white-to-all ratio (30\%, 50\%, 70\%). We trained different generators on the datasets with different representation ratio of groups, and analyse their ability to recover the distributions of both groups at the same rate. Similarly, for SVHN dataset, we darkened certain images (Fig. \ref{fig:snvh_snapshot}). We have considered samples brightness as a sensitive attribute, forming new group out of SVHN samples, divided by 2.

\subsection{Results on MNIST dataset}
\begin{figure*}[ht!]
    %\includegraphics[width=0.7\columnwidth]{./images/group_distrib.png} 
    %\begin{center} 
    \resizebox{\textwidth}{!}{\input{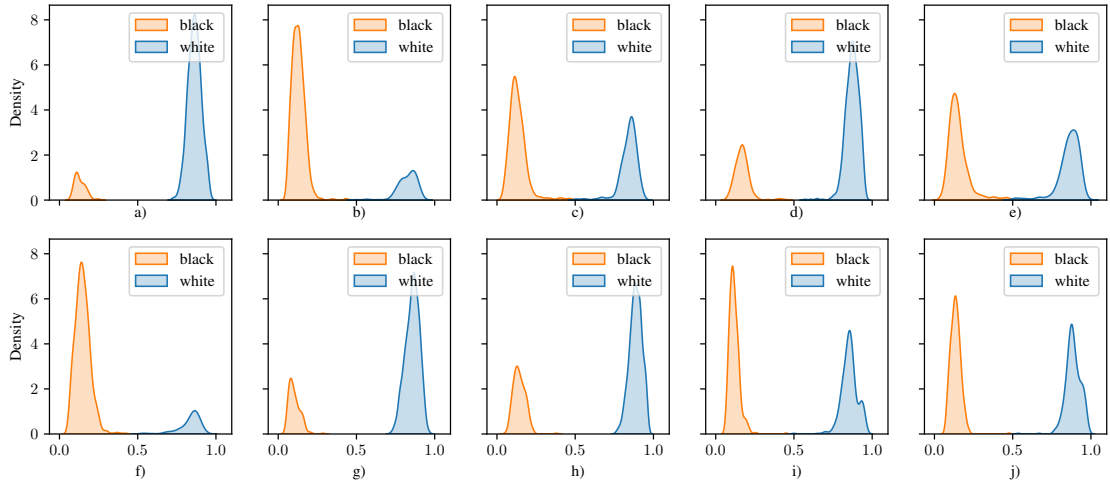}} 
    %\end{center}
    \caption{Distribution of digits generated by Stacked GAN trained on the modified MNIST dataset with balanced group representations: black digits ($mean < 0.5$); white digits ($mean > 0.5$). The generator evenly produces digits from both groups, but not in most cases (a, b, c, d, f, g).  }
    \label{fig:group_distrib_minist}
\end{figure*}

On MNIST dataset, we trained SGAN model with equal representation of different groups in the dataset (black and white digits) and analysed the distribution of groups inside the data generated by the generator. We used mean of the sample to decide which group each example belongs to: if the mean is greater than 0.5, then the sample is considered white and black otherwise. On overall, even thought groups were equally represented in the training dataset, the model was not able to generate white and black digits at equal rate. Figure \ref{fig:group_distrib_minist} presents joint distribution of generated digits means. We trained generators 10 times, sampling 100 images per each of 10 classes. Figure \ref{fig:group_distrib_minist} shows distribution of samples over all 10 experiments. 

For instance Fig. \ref{fig:group_distrib_minist}(a) shows that majority of samples are white, while in Fig. \ref{fig:group_distrib_minist}(b) majority are black. In most cases the group distribution is not well balanced. The desired case is the one in which the group distributions are the same (as in Fig. \ref{fig:group_distrib_minist}(j)). However throughout our experiments, this case has rarely occurred.      
 
\begin{figure*}[h!]
 
    %\includegraphics[width=0.7\columnwidth]{./images/group_distrib.png} 
    %\begin{center} 
    \resizebox{\textwidth}{!}{\input{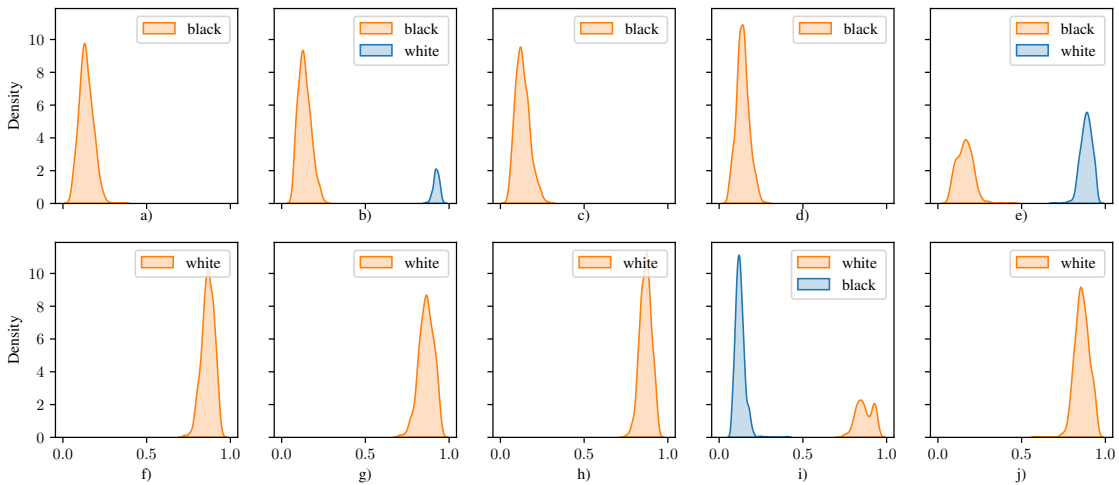}} 
    %\end{center}
    \caption{Distribution of digits generated using GAN trained on MNIST dataset with imbalance group representation. (a, b, c, d, e) dataset with  70\% of black digits vs 30\% of white digits. (f, g, h, i, j) dataset with 30\% of black digits vs 70\% of white digits. The generator mainly generates digits that belong to the well representation group.}
    \label{fig:group_distrib_minist_unbalance}
\end{figure*}

The discrepancy is even more serious if the groups' representations are not well balanced in training set. We designed different experiment settings where, the GANs model was trained with imbalance groups representations. In the first setting, the model was trained using a dataset containing 70\% of white digits and 30\% of black digits, and the opposite in the second setting.  In most cases, out of a thousand examples generated, most are from the well represented group. This behaviour shows groups are not treated equally by the model. Figure \ref{fig:group_distrib_minist_unbalance} shows the group distributions of examples generated using the trained model with the imbalanced group representation. 

\subsection{Results on SVHN dataset}
\begin{figure}[h!]
 
    %\includegraphics[width=0.7\columnwidth]{./images/group_distrib.png} 
    %\begin{center} 
    \resizebox{8.5cm}{!}{\input{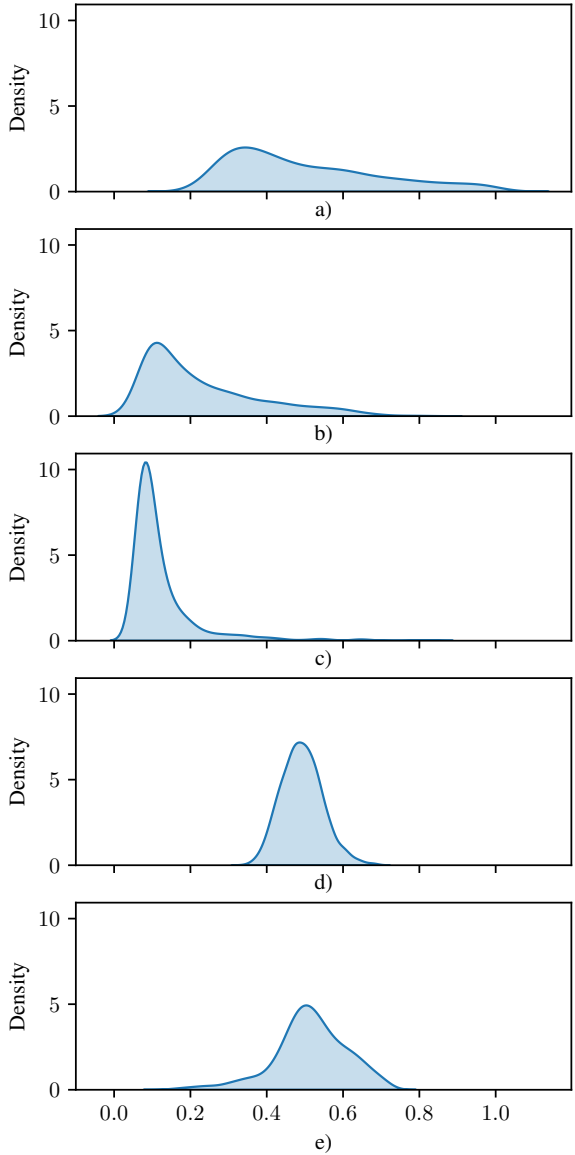}} 
    %\end{center}
    \caption{Results on the SVHN dataset. In most cases the generated examples are concentrated around one group. }
    \label{fig:group_distrib_svnh_unbalance}
\end{figure}

On SVHN datset we trained almost the same SGAN, as reported in the Fig. \ref{fig:group_distrib_svnh_unbalance}, we trained 5 different generators using the same dataset. Group representations were balanced within the training data. However, during the testing phase, the examples sampled from the trained generator had a different distribution. This suggests that the generator converged to a particular group during training (Fig. \ref{fig:group_distrib_svnh_unbalance}).       

\subsection{CelibA dataset}
\begin{figure*}
\centering
\begin{subfigure}[b]{0.49\textwidth}
\includegraphics[width=\textwidth]{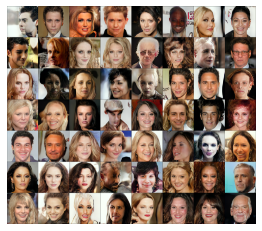}
%\caption{Pre processing memory graph}
\end{subfigure}
\begin{subfigure}[b]{0.49\textwidth}
\includegraphics[width=\textwidth]{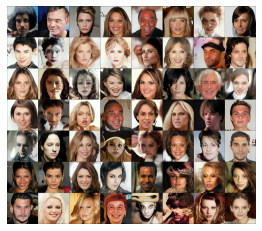}
%\caption{Post processing memory graph}
\end{subfigure}
\caption{Images generated using PGAN. Overall the generator predominately produces pictures of females (well represented groups) with lighter skins tones.}
\label{fig:celibA}
\end{figure*}

 CelibA dataset contains 61.4\% female faces versus 38.6\% male faces.
 We generated different images using Progressive Growing of GANs (PGAN) model, which is good in generating high resolution images \cite{karras2017progressive}, and it turns out that well-represented groups (female faces) are the most generated by the generator, while there is a lack of dark-skinned faces throughout several runs of the trained generator (Figure \ref{fig:celibA}). Moreover, most of the few black faces generated were of worse quality.

\section{Discussions}
In this section, we examine the impact that the inherent biases produced by GANs models can have when they are used to generate data for other downstream tasks, such as data augmentation, class imbalance problem and fair representation learning.  

\subsection{Class imbalance problem \& Data augmentation}
GANs model can be used to increase the size of training dataset when there is not enough data available. For example, dataset of credit card fraud will normally contains more records of normal transactions than fraudulent ones. Similarly, database of medical information will contains higher number of healthy patients than affected ones. For classification tasks, when different classes  are  not  uniformly distributed within the dataset, the accuracy can be significantly impacted negatively. Indeed, classifiers using this  kind  of  dataset  are  biased  towards  the  underrepresented classes. The good generative properties of GAN models gives the possibility to restore the dataset balance by generating examples from minority-class (oversampling).

Another useful application of GAN models for data augmentation is when due to privacy concerns, real dataset cannot be used directly. For example, medical records could contain a lot of sensitive information about patients that cannot be shared or are protected. GAN can be use in this case to generate synthetic dataset, that could be then used for a given task e.g classification.  

There are various applications of GANs in this way. Tanaka et. al \cite{tanaka2019data} analysed the performances of GANs model when used to balance the dataset and generate synthetic data for classification tasks, they showed that synthetic data generated by GAN can in some cases lead to better accuracy performance than real dataset. For instance, Giovanni et. al \cite{mariani2018bagan} used to restore the balance of an imbalanced dataset.  Other applications to balance dataset for Speech Emotion Recognition  \cite{chatziagapi2019data}.s

In this paper, we argue that despite the fact GAN provides interesting property for data augmentation, their inability to generate different groups that are uniformly distributed may rise fairness concerns. As the results of the previous section showed, GAN model can inherently tends to converge to prefer certain groups in the dataset and therefor generates a much larger number of instances from these groups. Thus, when used for dataset augmentation, this may leads to the creation of biased dataset, i.e. dataset in which different groups are not equally represented. It has been shown that classifiers trained using this kind of dataset may lead to unfair decisions \cite{mehrabi2019survey}.    

\subsection{Using GANs for fair representation learning}
Adversarial training has been used to learn "fair" representation of data. The goal of this approach is to mitigate unfairness by obstructing information concerning a protected attribute (Gender, age, religion etc...) as much as possible, while preserving other useful information \cite{feng2019learning}. The objective of the leaned representation is to achieve desired fairness notion such as statistical parity or individual fairness in downstream tasks as classification.
The potential problem with approaches that use adversarial learning to mitigate unfairness is the bias toward subgroups. In fact, when the adversarial model attempts to mitigate unfairness by removing all information about a given sensitive attribute, there is a risk that the generator will only converge to certain subgroups. For instance, if the sensitive is gender, with each group having subgroups (for example, pregnant women) then the generator may focus on the main subgroups (non-pregnant women) and then unfairness will shift away from a male-female problem to a pregnant and non-pregnant problem.

\section{Proposed solutions}
\label{sec:solution}
In this section, we discuss possible solutions that could help solve the problem presented in the previous sections.

\subsection{Using Conditional GANs}
One possible solution to solve the stated issue is to use Conditional GANs \cite{mirza2014conditional}. Instead of conditioning the generator and discriminator on the class label $y$, one can condition them with the groups label and necessary class label. The trained generator will then allows to generate samples conditioned by group label. e.g. generate images of imaginary actors of given gender (group) like male or female. This solution needs to be explored as quality of generated data. A limitation of this approach is that biases may shift to subgroups.  

\subsection{Using model ensemble}
Another interesting direction to explore is ensemble learning. The main idea behind ensemble model is to train different model and combine them to produce one model that outperform all of them. Boosting is an ensemble method technique widely for classification tasks. The idea is to create successive model during the training process, in which a newly created model attempts to correct the errors from the preview model. Models are added to the ensemble until errors are completely eliminated on training set or the maximum number of models is reached. 
 In section this subsection, we describe a technique works similarly to boosting algorithm, to build ensemble of generators that combined together,  could generate data with an equal representation rate of different groups.  During the training process, different generators can be created base on the previous ones and added to the ensemble. 
More precisely,  while creating new generator $G_i$ based on the set of the already existing generators $\textbf{G} = \{G_1, G_2, ..., G_{i-1}\}$, one should consider the set of corresponding samples generated by each existing generator $G_i$. i.e $\textbf{Q} = \{Q_1, Q_2, ..., Q_{i-1}\}$, where $Q_i$ is the set of samples generated by the generator $G_i$ i.e. $Q_i = \{x_1, x_2, ..., x_m\}, \; \forall j \in \{1, ..., i-1\}$; where $m$ is a hyper-parameter.

We can then add a regularization term to the loss function of generators, that will enforces the distance between the newly and previously generated samples. For a given set of fake samples $\textbf{F} = \{f_1, f_2, ..., f_m\}$, sampled by the generator at the current batch, the regularization term $\Theta$ can define as following:
\begin{equation}
\Theta =  \frac{\Phi(\textbf{Q})}{\sum_{Q_i \in \textbf{Q}}\sum_{x \in Q_i}  \: \sum_{f \in \textbf{F}}  \parallel x - f \parallel}
\end{equation}
Where $\Phi$ is a given function that depends on $\textbf{Q}$ used as normalization term.

Thus the new loss of a generator $G_i$ is define as follow:
\begin{equation}
    \mathcal{L}^{(G_i)} = - \mathbb{E}_z \mathrm{log}\;(1- \mathrm{D}(\mathrm{G}(z))) + \Theta_i
\end{equation}

%During the inference stage the goal is to generate samples from all the existing groups (and, possibly, sub-groups). In order to do it we should not use generators, which generating space is already spanned with previously generated samples ($\textbf{g} = \{g_1, g_2,..., g_p\}$), preferring to use generators, which may generate samples, different from the already generated ones.
During the testing phase, the objective is to generate samples from all existing groups (and possibly subgroups).  To do so, we should not use generators, whose generation space is already covered by already generated samples ($\textbf{g} = \{g_1, g_2,..., g_p\}$), but rather generators that can generate samples different from those already generated. So, given a probability density functions $\{P_1, P_2,..., P_n\}$ for all the generators $\{G_1, G_2,..., G_n\}$ and probability $\textbf{P}$ for the generator to be chosen, we have $\textbf{P}(G_i) < \textbf{P}(G_j)$ iff $\sum_{g \in \textbf{g}} P_i(g) > \sum_{g \in \textbf{g}} P_j(g)$.

Although we can not be sure about the form of the probability density function for the different generators, we may use a Normal distribution as an estimator for the function. In this case, $P_i = \mathcal{N}(E[X_i], \sqrt{Var[X_i]})$.

More specifically, probability for some generator $G_i$ to be chosen is given by the following equation:
\begin{equation}
\textbf{P}(G_i) = \frac{\displaystyle \prod_{f \in \textbf{f}}e^{1 - P_i(f)}}{\displaystyle \sum_{P \in \textbf{P}} \prod_{f \in \textbf{f}}e^{1 - P(f)}}
\end{equation}
In other word, $\textbf{P}(G_i)$ is defined as the softmax over probability that there is no point at $f$.
\section{Related work}
The existing works, which are similar to ours, are those that seek to solve the GAN mode collapse problem in GANs, by improving the generalization performance of the model, in order to generate diverse samples. For instance, Qi et. al \cite{mao2019mode} proposed a regularization method for mode seeking to address the mode collapse problem in CGANs. However, diversity of generated samples is different from the goal of fairly generating data from different groups. In other words, a diverse set of samples may also contain underrepresented groups samples. 

 Xu et. al \cite{xu2018fairgan} proposed FairGAN, a GAN architecture to generate "fair" data with respect to the statistic parity, i.e remove all the information about sensitive attribute from generated data, while retaining as much information as possible. In their architecture, they have included another discriminator, that distinguish whether the generated samples are from protected group or unprotected group. Although their work also focuses on fairness in GANs, the goal is different from ours in that we are trying to alleviate the problem of representational bias, rather than achieving a certain notion of fairness. However, in future work, we will conduct comprehensive empirically comparisons of our boosting approach and FairGAN approach in terms of accuracy-fairness tradeoff.     

\section*{Conclusion}
In this paper, we highlighted fairness concern of GANs model. We have shown that when different groups are present in the dataset, the GAN model is not able to generate them homogeneously during the test phase, which can lead to representation bias. We discussed existing applications of GANs and argued that techniques using GANs to generate data for downstream tasks should consider this issue in order to avoid generating biased data, inherently introduced by GANs.  We proposed a boosting strategy that can help to solve this issue. This strategy will be further explored in our future work and will be compared to other approaches in terms of fairness and accuracy.

\bibliographystyle{IEEEtran}
\bibliography{main.bib}

\begin{thebibliography}{10}
\providecommand{\url}[1]{#1}
\csname url@samestyle\endcsname
\providecommand{\newblock}{\relax}
\providecommand{\bibinfo}[2]{#2}
\providecommand{\BIBentrySTDinterwordspacing}{\spaceskip=0pt\relax}
\providecommand{\BIBentryALTinterwordstretchfactor}{4}
\providecommand{\BIBentryALTinterwordspacing}{\spaceskip=\fontdimen2\font plus
\BIBentryALTinterwordstretchfactor\fontdimen3\font minus
  \fontdimen4\font\relax}
\providecommand{\BIBforeignlanguage}[2]{{%
\expandafter\ifx\csname l@#1\endcsname\relax
\typeout{** WARNING: IEEEtran.bst: No hyphenation pattern has been}%
\typeout{** loaded for the language `#1'. Using the pattern for}%
\typeout{** the default language instead.}%
\else
\language=\csname l@#1\endcsname
\fi
#2}}
\providecommand{\BIBdecl}{\relax}
\BIBdecl

\bibitem{liu2017survey}
W.~Liu, Z.~Wang, X.~Liu, N.~Zeng, Y.~Liu, and F.~E. Alsaadi, ``A survey of deep
  neural network architectures and their applications,'' \emph{Neurocomputing},
  vol. 234, pp. 11--26, 2017.

\bibitem{vaswani2017attention}
A.~Vaswani, N.~Shazeer, N.~Parmar, J.~Uszkoreit, L.~Jones, A.~N. Gomez,
  L.~Kaiser, and I.~Polosukhin, ``Attention is all you need,'' \emph{arXiv
  preprint arXiv:1706.03762}, 2017.

\bibitem{rivera2020anomaly}
A.~R. Rivera, A.~Khan, I.~E.~I. Bekkouch, and T.~S. Sheikh, ``Anomaly detection
  based on zero-shot outlier synthesis and hierarchical feature distillation,''
  \emph{IEEE Transactions on Neural Networks and Learning Systems}, 2020.

\bibitem{gavrilin2019across}
Y.~Gavrilin and A.~Khan, ``Across-sensor feature learning for energy-efficient
  activity recognition on mobile devices,'' in \emph{2019 International Joint
  Conference on Neural Networks (IJCNN)}.\hskip 1em plus 0.5em minus
  0.4em\relax IEEE, 2019, pp. 1--7.

\bibitem{sozykin2018multi}
K.~Sozykin, S.~Protasov, A.~Khan, R.~Hussain, and J.~Lee, ``Multi-label
  class-imbalanced action recognition in hockey videos via 3d convolutional
  neural networks,'' in \emph{2018 19th IEEE/ACIS International Conference on
  Software Engineering, Artificial Intelligence, Networking and
  Parallel/Distributed Computing (SNPD)}.\hskip 1em plus 0.5em minus
  0.4em\relax IEEE, 2018, pp. 146--151.

\bibitem{khan2010accelerometer}
A.~M. Khan, Y.-K. Lee, S.~Lee, and T.-S. Kim, ``Accelerometer’s position
  independent physical activity recognition system for long-term activity
  monitoring in the elderly,'' \emph{Medical \& biological engineering \&
  computing}, vol.~48, no.~12, pp. 1271--1279, 2010.

\bibitem{protasov2018using}
S.~Protasov, A.~M. Khan, K.~Sozykin, and M.~Ahmad, ``Using deep features for
  video scene detection and annotation,'' \emph{Signal, Image and Video
  Processing}, vol.~12, no.~5, pp. 991--999, 2018.

\bibitem{bengio2013representation}
Y.~Bengio, A.~Courville, and P.~Vincent, ``Representation learning: A review
  and new perspectives,'' \emph{IEEE transactions on pattern analysis and
  machine intelligence}, vol.~35, no.~8, pp. 1798--1828, 2013.

\bibitem{khan2020post}
A.~Khan and K.~Fraz, ``Post-training iterative hierarchical data augmentation
  for deep networks,'' \emph{Advances in Neural Information Processing
  Systems}, vol.~33, 2020.

\bibitem{kingma2013auto}
D.~P. Kingma and M.~Welling, ``Auto-encoding variational bayes,'' \emph{arXiv
  preprint arXiv:1312.6114}, 2013.

\bibitem{rezende2014stochastic}
D.~J. Rezende, S.~Mohamed, and D.~Wierstra, ``Stochastic backpropagation and
  approximate inference in deep generative models,'' in \emph{International
  conference on machine learning}.\hskip 1em plus 0.5em minus 0.4em\relax PMLR,
  2014, pp. 1278--1286.

\bibitem{goodfellow2014generative}
I.~J. Goodfellow, J.~Pouget-Abadie, M.~Mirza, B.~Xu, D.~Warde-Farley, S.~Ozair,
  A.~Courville, and Y.~Bengio, ``Generative adversarial networks,'' \emph{arXiv
  preprint arXiv:1406.2661}, 2014.

\bibitem{karras2017progressive}
T.~Karras, T.~Aila, S.~Laine, and J.~Lehtinen, ``Progressive growing of gans
  for improved quality, stability, and variation,'' \emph{arXiv preprint
  arXiv:1710.10196}, 2017.

\bibitem{guo2018long}
J.~Guo, S.~Lu, H.~Cai, W.~Zhang, Y.~Yu, and J.~Wang, ``Long text generation via
  adversarial training with leaked information,'' in \emph{Proceedings of the
  AAAI Conference on Artificial Intelligence}, vol.~32, no.~1, 2018.

\bibitem{mogren2016c}
O.~Mogren, ``C-rnn-gan: Continuous recurrent neural networks with adversarial
  training,'' \emph{arXiv preprint arXiv:1611.09904}, 2016.

\bibitem{esteban2017real}
C.~Esteban, S.~L. Hyland, and G.~R{\"a}tsch, ``Real-valued (medical) time
  series generation with recurrent conditional gans,'' \emph{arXiv preprint
  arXiv:1706.02633}, 2017.

\bibitem{madras2018learning}
D.~Madras, E.~Creager, T.~Pitassi, and R.~Zemel, ``Learning adversarially fair
  and transferable representations,'' in \emph{International Conference on
  Machine Learning}.\hskip 1em plus 0.5em minus 0.4em\relax PMLR, 2018, pp.
  3384--3393.

\bibitem{dwork2012fairness}
C.~Dwork, M.~Hardt, T.~Pitassi, O.~Reingold, and R.~Zemel, ``Fairness through
  awareness,'' in \emph{Proceedings of the 3rd innovations in theoretical
  computer science conference}, 2012, pp. 214--226.

\bibitem{mehrabi2019survey}
N.~Mehrabi, F.~Morstatter, N.~Saxena, K.~Lerman, and A.~Galstyan, ``A survey on
  bias and fairness in machine learning,'' \emph{arXiv preprint
  arXiv:1908.09635}, 2019.

\bibitem{arjovsky2017wasserstein}
M.~Arjovsky, S.~Chintala, and L.~Bottou, ``Wasserstein generative adversarial
  networks,'' in \emph{International conference on machine learning}.\hskip 1em
  plus 0.5em minus 0.4em\relax PMLR, 2017, pp. 214--223.

\bibitem{radford2015unsupervised}
A.~Radford, L.~Metz, and S.~Chintala, ``Unsupervised representation learning
  with deep convolutional generative adversarial networks,'' \emph{arXiv
  preprint arXiv:1511.06434}, 2015.

\bibitem{mirza2014conditional}
M.~Mirza and S.~Osindero, ``Conditional generative adversarial nets,''
  \emph{arXiv preprint arXiv:1411.1784}, 2014.

\bibitem{odena2017conditional}
A.~Odena, C.~Olah, and J.~Shlens, ``Conditional image synthesis with auxiliary
  classifier gans,'' in \emph{International conference on machine
  learning}.\hskip 1em plus 0.5em minus 0.4em\relax PMLR, 2017, pp. 2642--2651.

\bibitem{huang2017stacked}
X.~Huang, Y.~Li, O.~Poursaeed, J.~Hopcroft, and S.~Belongie, ``Stacked
  generative adversarial networks,'' in \emph{Proceedings of the IEEE
  conference on computer vision and pattern recognition}, 2017, pp. 5077--5086.

\bibitem{lecun1998gradient}
Y.~LeCun, L.~Bottou, Y.~Bengio, and P.~Haffner, ``Gradient-based learning
  applied to document recognition,'' \emph{Proceedings of the IEEE}, vol.~86,
  no.~11, pp. 2278--2324, 1998.

\bibitem{netzer2011reading}
Y.~Netzer, T.~Wang, A.~Coates, A.~Bissacco, B.~Wu, and A.~Y. Ng, ``Reading
  digits in natural images with unsupervised feature learning,'' 2011.

\bibitem{tanaka2019data}
F.~H. K. d.~S. Tanaka and C.~Aranha, ``Data augmentation using gans,''
  \emph{arXiv preprint arXiv:1904.09135}, 2019.

\bibitem{mariani2018bagan}
G.~Mariani, F.~Scheidegger, R.~Istrate, C.~Bekas, and C.~Malossi, ``Bagan: Data
  augmentation with balancing gan,'' \emph{arXiv preprint arXiv:1803.09655},
  2018.

\bibitem{chatziagapi2019data}
A.~Chatziagapi, G.~Paraskevopoulos, D.~Sgouropoulos, G.~Pantazopoulos,
  M.~Nikandrou, T.~Giannakopoulos, A.~Katsamanis, A.~Potamianos, and
  S.~Narayanan, ``Data augmentation using gans for speech emotion
  recognition.'' in \emph{Interspeech}, 2019, pp. 171--175.

\bibitem{feng2019learning}
R.~Feng, Y.~Yang, Y.~Lyu, C.~Tan, Y.~Sun, and C.~Wang, ``Learning fair
  representations via an adversarial framework,'' \emph{arXiv preprint
  arXiv:1904.13341}, 2019.

\bibitem{mao2019mode}
Q.~Mao, H.-Y. Lee, H.-Y. Tseng, S.~Ma, and M.-H. Yang, ``Mode seeking
  generative adversarial networks for diverse image synthesis,'' in
  \emph{Proceedings of the IEEE/CVF Conference on Computer Vision and Pattern
  Recognition}, 2019, pp. 1429--1437.

\bibitem{xu2018fairgan}
D.~Xu, S.~Yuan, L.~Zhang, and X.~Wu, ``Fairgan: Fairness-aware generative
  adversarial networks,'' in \emph{2018 IEEE International Conference on Big
  Data (Big Data)}.\hskip 1em plus 0.5em minus 0.4em\relax IEEE, 2018, pp.
  570--575.

\end{thebibliography}
 
\end{document}